\documentclass[runningheads]{llncs}
\usepackage[T1]{fontenc}
\usepackage{graphicx}
\usepackage{ upgreek }
\usepackage{amsfonts}
\usepackage{xspace}
\usepackage{float}
\usepackage{wrapfig}
\usepackage{makecell}
\usepackage{multirow}
\usepackage{graphicx}
\usepackage[table,xcdraw]{xcolor}
\usepackage{amsmath}
\usepackage{hyperref}

\def\model{SPI-CorrNet\xspace}  %Uncertain-SpICE

\newcommand{\myvector}[1]{\mathbf{\lowercase{#1}}}
\newcommand{\set}[1]{\mathcal{#1}}
\newcommand{\realdim}[1]{\mathbb{#1}}
\newcommand{\mymatrix}[1]{\mathbf{\uppercase{#1}}}
\begin{document}
\title{Probabilistic 3D Correspondence Prediction from Sparse Unsegmented Images}

\titlerunning{\model}
\author{Krithika Iyer\inst{1,2} \and
Shireen Y. Elhabian\inst{1,2}}
% \author{*}
%
\authorrunning{Iyer and Elhabian}
\institute{Scientific Computing and Imaging Institute, University of Utah, UT, USA \and
Kahlert School of Computing, University of Utah, UT, USA \\
\email{krithika.iyer@utah.edu } \email{shireen@sci.utah.edu}}
\maketitle              % typeset the header of the contribution

\vspace{-8mm}
\begin{abstract}
% intro about SSM and the drawbacks of tradiational approaches
The study of physiology demonstrates that the form (shape) of anatomical structures dictates their functions, and analyzing the form of anatomies plays a crucial role in clinical research. Statistical shape modeling (SSM) is a widely used tool for quantitative analysis of forms of anatomies, aiding in characterizing and identifying differences within a population of subjects. Despite its utility, the conventional SSM construction pipeline is often complex and time-consuming. Additionally, reliance on linearity assumptions further limits the model from capturing clinically relevant variations. %, collectively impeding its widespread adoption. 
% advances in deep learning methods and the research gap
Recent advancements in deep learning solutions enable the direct inference of SSM from unsegmented medical images, streamlining the process and improving accessibility. However, the new methods of SSM from images do not adequately account for situations where the imaging data quality is poor or where only sparse information is available. Moreover, quantifying aleatoric uncertainty, which represents inherent data variability, is crucial in deploying deep learning for clinical tasks to ensure reliable model predictions and robust decision-making, especially in challenging imaging conditions.
% brief description of proposed model
Therefore, we propose \model, a unified model that predicts 3D correspondences from sparse imaging data. It leverages a teacher network to regularize feature learning and quantifies data-dependent aleatoric uncertainty by adapting the network to predict intrinsic input variances. Experiments on the LGE MRI left atrium dataset and Abdomen CT-1K liver datasets demonstrate that our technique enhances the accuracy and robustness of sparse image-driven SSM.

\end{abstract}
% \vspace{-10.5mm}
\vspace{-4.5mm}
\section{Introduction}
% \vspace{-3.5mm}
Understanding morphological variations influenced by pathology, gender, and age is crucial for personalized treatment strategies in precision medicine, facilitating fast diagnosis and treatment \cite{singh2020evaluation,dai2020statistical}. Statistical Shape Modeling (SSM) is pivotal in medical image analysis, enabling the identification of morphological variations and quantitative assessment of geometric variability across populations. SSM applications include lesion screening, surgical planning, implant design \cite{li2024back}, and studying disease progression \cite{peirlinck2021precision}.

SSM parameterizes shapes into numerical vectors for statistical analysis. Methods for shape parameterization include implicit representations (e.g., deformation fields \cite{durrleman2014morphometry}, level set methods \cite{samson2000level}) and explicit representations such as an ordered set of landmarks or {\textit{correspondence points}} (aka point distribution models, PDMs), which describe anatomically equivalent points across samples. PDMs are favored for their ease of interpretation, computational efficiency, and noise tolerance.

Correspondences in SSM can be established manually or automatically by minimizing objective functions \cite{cates2008particle,cates2017shapeworks,davies2002learning,styner2006framework}. Traditional methods are often complex, computationally demanding, and require anatomical expertise, making them inadequate for large datasets. Deep learning models \cite{bhalodia2024deepssm} simplify the process by training directly from unsegmented images, but they still depend on computationally derived PDMs for supervision, which can bias and limit the models.

High-quality medical images are crucial for accurate shape models, but capturing dense, high-resolution images is challenging and costly. Sparse imaging, with limited data points or slices, arises from acquisition time constraints, patient comfort, radiation dose considerations \cite{schultz2020risk}, or technical limitations \cite{hollingsworth2015reducing}. Enhancing image resolution through post-acquisition resampling can reduce diagnostic accuracy, making it essential to develop models that extract meaningful information from sparse imaging content. Additionally, processing dense, high-resolution imaging data requires significant computational resources and time. Sparse imaging reduces data size and complexity, facilitating efficient shape modeling \cite{shieh2019spare} and benefiting real-time applications like intraoperative guidance or rapid diagnostic assessments.

Uncertainty quantification is crucial in clinical applications to avoid overconfident and unreliable estimates. Aleatoric uncertainty (data dependent), inherent in sparse imaging due to noise and variability, must be accounted for in SSM methods to produce robust and reliable shape models. Quantifying uncertainty informs clinicians about the potential risks and limitations of the model's outcomes.

To address these challenges, we propose the \textbf{\underline{Sp}}arse \textbf{\underline{I}}mage-base probabilistic \textbf{\underline{Corr}}espondence \textbf{\underline{Net}}work (\model) for inferring 3D correspondences from sparse, unsegmented medical images with incorporated aleatoric uncertainty estimates. Our model leverages the student-teacher framework from SCorP \cite{iyer2024scorp} to learn a shape prior, which regularizes correspondence prediction and ensures anatomical accuracy in reconstructed shapes. Notably, our approach does not require ground PDMs for supervision, effectively managing variability and noise in sparse imaging data.
\vspace{-4mm}
\section{Related Work}
\vspace{-2.5mm}
Various methods are used in shape analysis to establish correspondences. Non-optimized methods involve manual annotation and warping landmarks using registration techniques \cite{paulsen2002building,heitz2005statistical}, producing inconsistent results for larger populations. Parametric methods like SPHARM-PDM \cite{styner2006framework} use fixed geometrical bases for pairwise correspondences but struggle with complex shapes. Group-wise non-parametric approaches, such as particle-based shape modeling (PSM) \cite{cates2008particle,cates2017shapeworks} and minimum description length (MDL) \cite{davies2002learning}, consider cohort variability and optimize data-driven objectives. Deep learning models simplify conventional SSM pipelines by performing supervised correspondence prediction directly from unsegmented images (TL-DeepSSM and DeepSSM \cite{bhalodia2024deepssm}).

In clinical applications, uncertainty quantification is essential for evaluating tool reliability. Recent advancements include aleatoric (data-dependent) and epistemic (model-dependent) uncertainty estimation. Aleatoric uncertainty is modeled by a probability distribution over the outputs, while epistemic uncertainty is captured using Bayesian neural networks \cite{gal2016uncertainty} or ensemble methods. Uncertain DeepSSM \cite{adams2020uncertain} includes both types of uncertainty estimation but, it relies on a shape prior in the form of a supervised latent encoding pre-computed using principal component analysis (PCA). Similarly, other models were proposed for probabilistic 2D surface reconstruction using PCA scores as a prior \cite{tothova2018uncertainty}, which was also extended to probabilistic 3D surface reconstruction from sparse 2D images \cite{tothova2020probabilistic}. Although these approaches provide shape segmentation with aleatoric uncertainty measures, they do not offer a shape representation readily usable for population-level statistical analysis. VIB-DeepSSM \cite{adams2022images} relaxes the PCA assumption using a variational information bottleneck (VIB) \cite{alemi2016deepvib} for latent encoding learning, improving aleatoric uncertainty estimation and generalization, but cannot measure epistemic uncertainty fully. The fully Bayesian BVIB-DeepSSM \cite{adams2023fully} addresses this by quantifying both uncertainties and predicting probabilistic shapes from images. But BVIV-DeepSSM and VIB-DeepSSM continue to rely on established PDM for supervision. 

Recent models like FlowSSM \cite{ludke2022landmark}, Point2SSM \cite{adams2023point2ssm}, Mesh2SSM \cite{iyer2023mesh2ssm}, and SCorP \cite{iyer2024scorp} predict correspondences from various data modalities without requiring PDMs for supervision. Particularly, SCorP \cite{iyer2024scorp} incorporates a shape prior learned from surface meshes in a student-teacher framework for regularizing feature learning from images without supervised PDM loss, but these approaches lack uncertainty estimation.

Existing methods for probabilistic correspondence prediction face limitations such as imposing a linear relationship between latent and output spaces and reliance on predefined PDMs for training. Additionally, sparse data is not utilized for building shape models. The proposed framework aims to directly predict correspondences and estimate uncertainty from sparse, unsegmented images without predefined PDMs during training.
\vspace{-4mm}
\section{Background}
\vspace{-2mm}
Our work builds upon the model SCorP \cite{iyer2024scorp}. This section provides an overview of SCorP, setting the stage for our proposed method in Section \ref{methods}.\\
\textbf{SCorP Overview}\\
Consider a training dataset \(\set{S} = \{S_1, S_2, \ldots, S_N\}\) comprising \(N\) aligned surface meshes, and their corresponding aligned volumetric images, \(\set{I} = \{\mymatrix{I}_1, \mymatrix{I}_2, \ldots, \mymatrix{I}_N\}\). Each surface mesh \(S_j = (\set{V}_j, \set{E}_j)\) consists of vertices \(\set{V}_j\) and edge connectivity \(\set{E}_j\). The primary objective of SCorP is to establish a shape prior (\textit{teacher}) by predicting a set of \(M\) correspondence points \(\set{C}_j^{S} = \{\mathbf{c}_{j(1)}, \mathbf{c}_{j(2)}, \ldots, \mathbf{c}_{j(M)}\}\) with \(\mathbf{c}_{j(m)} \in \realdim{R}^3\), which accurately represent the anatomy described by surface mesh \(S_j\). This shape prior is then used to guide the image encoder (\textit{student}) in learning image representation \(\myvector{z}^I_j\) conducive to predicting a corresponding set of points \(\set{C}_j^{I} = \{\mathbf{c}_{j(1)}, \mathbf{c}_{j(2)}, \ldots, \mathbf{c}_{j(M)}\}\) directly from the associated image \(\mymatrix{I}_j\). SCorP's model architecture Fig~\ref{fig:model_arch}.A

\vspace{-4mm}
\section{Proposed Model} \label{methods}
\begin{figure}
    \vspace{-5mm}
    \centering
    \includegraphics[width=\textwidth]{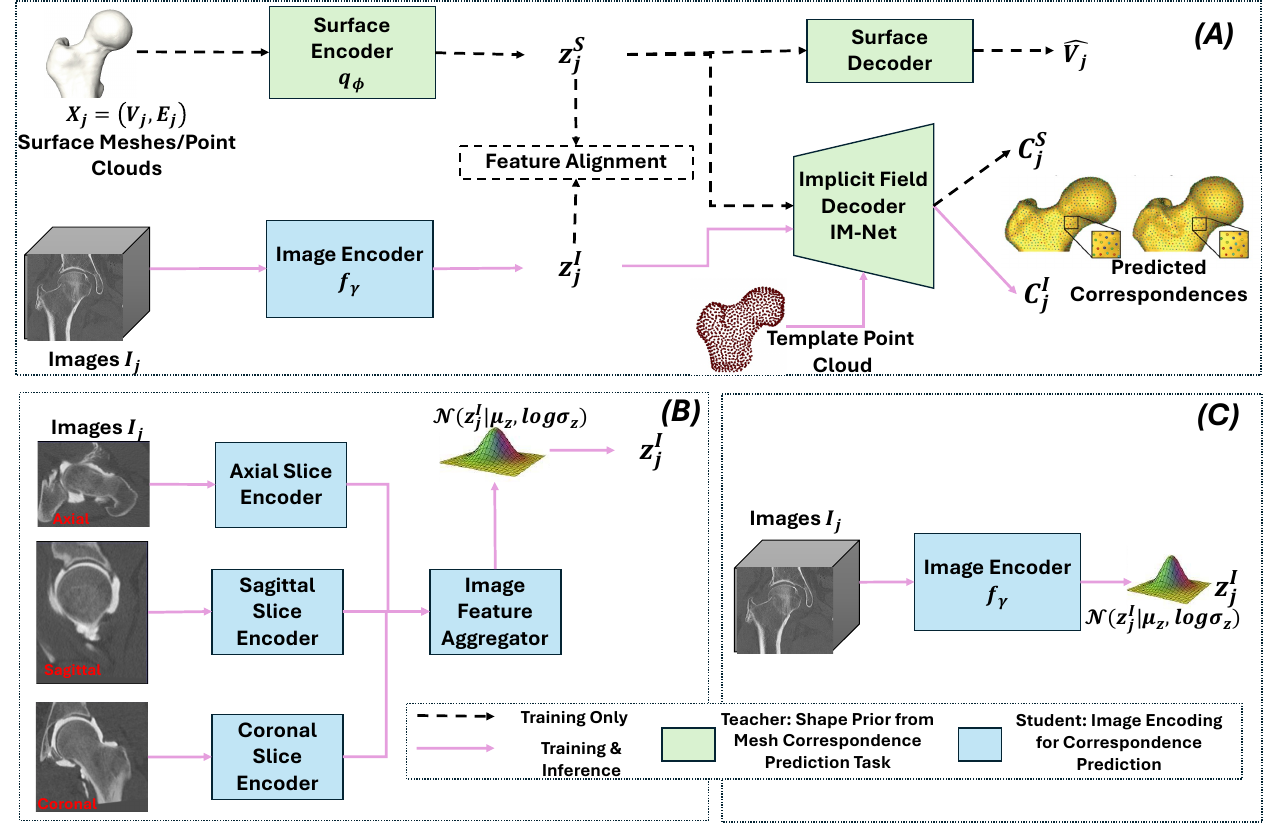}
    \vspace{-8mm}
    \caption{\textbf{Architecture:} (A) Base model SCorP \cite{iyer2024scorp} with the teacher (surface autoencoder and IM-NET decoder) and student (image encoder) network. Proposed modifications for \textbf{\model}: student network to handle (B) orthogonal image slices with a probabilistic image encoder and (C) full images with a probabilistic image encoder.}
    \label{fig:model_arch}
    \vspace{-7mm}
\end{figure}

\vspace{-2mm}
Our goal is to predict probabilistic 3D correspondence from sparse imaging. Specifically, we consider an input set of images \(\set{I} = \{\mymatrix{I}_1, \mymatrix{I}_2, \ldots, \mymatrix{I}_N\}\), where each image \(\mymatrix{I}_j\) consists of axial, coronal, and sagittal orthogonal slices (\(\mymatrix{I}_j = \{\mymatrix{I}_j^{AX}, \mymatrix{I}_j^{SG}, \mymatrix{I}_j^{CR}\}\)). To accommodate sparse imaging, we modify the student branch image encoder using three separate 2D CNNs to extract features from the three orthogonal slices. These features are concatenated and passed through an image feature aggregator network (fully connected layers) to predict a single latent vector \(\myvector{z}_j^I\) representing the entire sample. This modification ensures that we can adopt the student-teacher framework and use the same training strategy proposed by SCorP \cite{iyer2024scorp}. The modified student branch for \model is shown in Fig~\ref{fig:model_arch}.B.

Similar to SCorP, \model comprises a teacher network and a student network. The teacher network includes a surface autoencoder and an implicit field decoder. The surface autoencoder learns a low-dimensional \(\myvector{z}_j^S\), permutation invariant representation of each surface mesh using dynamic graph convolution with EdgeConv blocks \cite{wang2019dynamic}, while the IM-NET decoder \cite{chen2019net} uses this latent representation \(\myvector{z}_j^S\) to predict a set of correspondence points \(\set{C}_j^{S}\), ensuring consistency across the dataset by transforming a template point cloud to match each sample. The student network consists of the modified image encoder branch that learns a compact representation \(\myvector{z}_j^I\) for sparse image slices, capable of predicting a set of correspondence points \(\set{C}_j^{I}\) guided by the shape prior from the teacher network.

\vspace{-4mm}
\paragraph{Training} occurs in three phases similar to SCorP: (a) surface branch training to develop the shape prior where surface autoencoder and implicit decoder are jointly trained to minimize loss
\vspace{-2mm}
\begin{equation}\label{mesh_branch_loss}
    \mathcal{L}_{S} =  \sum_{j=1}^N \left[\mathcal{L}_{CD}(\set{V}_j,\set{C}_j^{S}) +  \alpha \mathcal{L}_{MSE}(\set{V}_j,\hat{\set{V}_j}) \right]
\end{equation}
\vspace{-0.5mm} 
where \(\hat{\set{V}_j}\) are the reconstructed vertex locations and \(\alpha\) is the weighting parameter, (b) image branch embedding alignment to align image encoder features with those of the surface encoder with loss function
\vspace{-2mm}
\begin{equation}\label{image_loss_phase2}
\mathcal{L}_{\text{EA}} = \frac{1}{N} \sum_{j=1}^{N} \left[ {| q_{\phi}(\myvector{z}^S_j|S_j) - {f}_{\gamma} (\myvector{z}^I_j|\mymatrix{I}_j)|}^2 \right]
\end{equation}
\vspace{-0.5mm}
and (c) image branch prediction refinement to improve correspondence prediction accuracy with loss combination of \(\mathcal{L}_{\text{PR}} + \mathcal{L}_{\text{EA}}\) where
\vspace{-2mm}
\begin{equation}\label{image_loss_phase3}
\mathcal{L}_{\text{PR}} =\sum_{j=1}^N \mathcal{L}_{L_2 CD}(\set{V}_j,{\set{C}_j^{I}})
\end{equation}
\vspace{-0.5mm}

\vspace{-7mm}
\paragraph{\textbf{Uncertainty Estimation:}} To incorporate probabilistic correspondence prediction for aleatoric uncertainty estimation, we propose making the student branch image encoder probabilistic. The encoder, \({f}_{\gamma}\), (comprising 3D convolutional and densely connected layers for full images Fig~\ref{fig:model_arch}.C and separate image slice encoder and image feature aggregator for sparse images Fig~\ref{fig:model_arch}.B) maps the input image \(\mymatrix{I}_j\) to a Gaussian latent distribution: \( \mathcal{N}(\myvector{z}^I|\mu_{\myvector{z}^I},\operatorname{log}\sigma_{\myvector{z}^I})\). Posterior samples \(\myvector{z}_j^I\) are acquired from this predicted latent distribution using the reparameterization trick to enable gradient calculation. This modification captures aleatoric uncertainty as the variance of the \(p(\set{C}_j^{I}|\myvector{z}^I)\) distribution, computed by sampling multiple latent encodings from \(\mathcal{N}(\myvector{z}^I|\mu_{\myvector{z}^I},\operatorname{log}\sigma_{\myvector{z}^I})\) and passing them through the implicit decoder to get a sampled distribution of predictions. A Gaussian distribution is estimated from these samples: \(\mathcal{N}(\set{C}_j^{I}|\mu, \operatorname{log}\sigma)\). The estimated \(\sigma\) captures the aleatoric uncertainty.

\vspace{-4mm}
\section{Dataset and Evaluation}
\vspace{-2.5mm}
\subsection{Datasets}
We selected the left atrium and liver datasets for our experiments due to their highly variable shapes. The \textbf{left atrium (LA)} dataset consists of 923 anonymized Late Gadolinium Enhancement (LGE) MRIs from distinct patients, manually segmented by cardiovascular medicine experts. Post-segmentation, the images were cropped around the region of interest. The \textbf{AbdomenCT-1K liver} dataset \cite{Ma-2021-AbdomenCT-1K} includes 1132 3D CT scans and their corresponding liver segmentations. After visually assessing the quality of the images and segmentations, we selected 833 samples. These images were aligned and cropped around the region of interest. We randomly split both datasets into training, validation, and test sets as \(80\%/10\%/10\%\). More details about the datasets, hyperparameters of the models, and training details are provided in the supplementary material.
\vspace{-2.5mm}
\subsection{Metrics}
\textbf{Chamfer Distance (CD)}: Measures the average bidirectional distance between points in two sets (\(\set{V}_j\) and \(\set{C}_j^{I}\)), assessing dissimilarity between them.
\textbf{Point-to-Mesh Distance (P2M)}: Calculates the sum of point-to-mesh face distance and face-to-point distance for the predicted correspondences (\(\set{C}_j^{I}\)) and the mesh faces defined by vertices and edges (\(\set{V}_j, \set{E}_j\)).
\textbf{Surface-to-Surface (S2S) Distance}: Measured between the original surface mesh and the generated mesh from predicted correspondences. To obtain the reconstructed mesh, correspondences are mapped to the mean shape, and the warp between the points is applied to its mesh.
\textbf{SSM Metrics} used to evaluate correspondence \cite{munsell2008evaluating}:
\textbf{Compactness}: Represents the training data distribution with minimal parameters, measured by the number of PCA modes needed to capture 95\% of the variation in correspondence points.
\textbf{Generalization}: Evaluates how well the SSM extrapolates from training to unseen examples, gauged by the reconstruction error (L2) between held-out and training SSM-reconstructed correspondence points.
\textbf{Specificity}: Measures the SSM's ability to generate valid instances of the trained shape class, quantified by the average distance between sampled SSM correspondences and the nearest existing training correspondences.
\textbf{Aleatoric Uncertainty}: Reflects inherent data noise and variability, expected to correlate with P2M error (high Pearson $r$). Aids in out-of-distribution detection, indicating model reliability.

\vspace{-4mm}
\section{Results}
\vspace{-2.5mm}
\begin{figure}
    \vspace{-5mm}
    \centering
    \includegraphics[width=\textwidth]{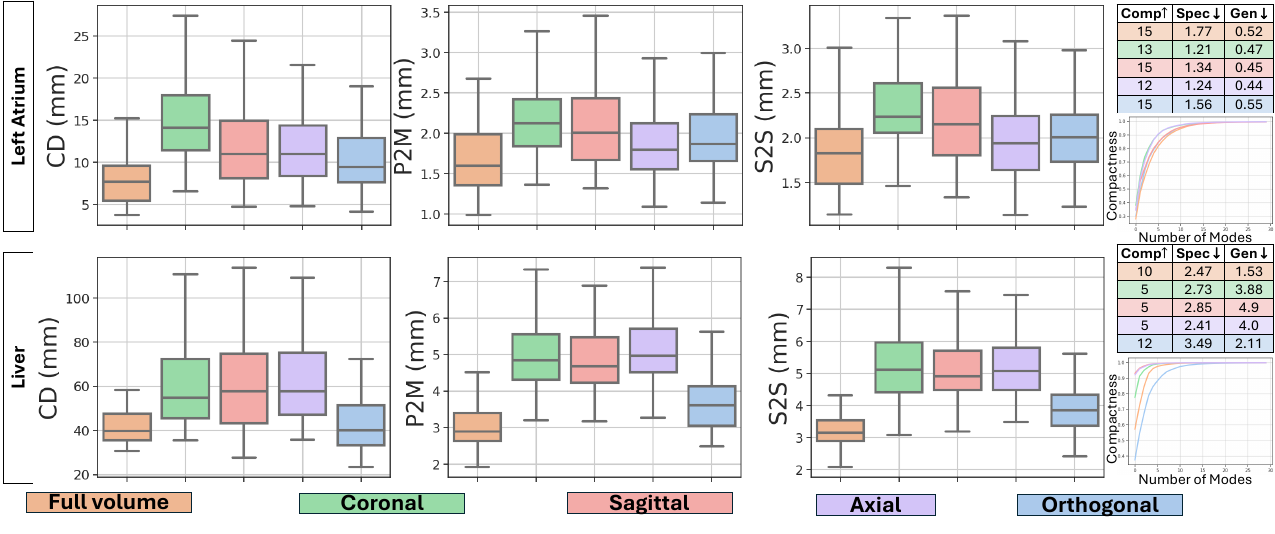}
    \vspace{-8mm}
    \caption{\textbf{Performance metrics:} Boxplots show performance metrics for held-out test samples from the LA and liver datasets. Compactness plots show cumulative variation captured by PCA modes. Comp = Compactness, Spec = Specificity, Gen = Generalization.}
    \label{fig:metrics}
     \vspace{-5mm}
\end{figure}

\begin{figure}
     % \vspace{mm}
    \centering
    \includegraphics[width=\textwidth]{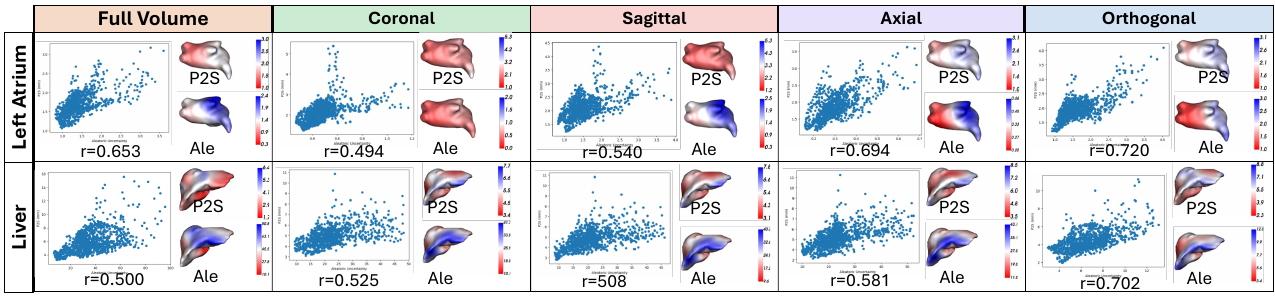}
    \vspace{-9mm}
    \caption{\textbf{Uncertainty Calibration:} Scatter plots and Pearson R coefficients show the correlation between estimated uncertainty and P2S error across test sets. Heatmaps on a representative mesh display average P2S error and aleatoric uncertainty, highlighting spatial correlation.}
    \label{fig:p2s_ale}
     \vspace{-7mm}
\end{figure}

We compare five \model variants using different input types: full volume (like SCorP), sparse images (orthogonal slices: axial, coronal, sagittal), and individual slices (axial, sagittal, coronal). This comparison identifies the most effective approach for probabilistic correspondence prediction. As shown in Fig~\ref{fig:metrics}, the full volume model outperforms others across CD, P2M, and S2S metrics. However, the proposed orthogonal slices model demonstrates competitive performance and is the second-best for both datasets. Notably, the axial slice model performs similarly to the three-slice model for the LA dataset, likely due to its effective capture of essential LA shape features such as length and appendage. Additionally, training the orthogonal slices model is 1.5x faster than the full volume model which highlights the utility of using sparse imaging for SSM applications. 

All models exhibit similar performance in SSM metrics (generalization, specificity, and compactness) as shown in Fig~\ref{fig:metrics}, indicating they capture significant shape variability while maintaining high fidelity to the original shapes. The compactness plots suggest an efficient representation of population variance with fewer PCA modes. Specificity and generalization metrics confirm that \model generates valid instances and effectively extrapolates to unseen data, regardless of input type. The full-volume model shows the best SSM metrics for the liver dataset, likely due to higher image quality and greater variation within the dataset.

\begin{figure}
    \vspace{-5mm}
    \centering
    \includegraphics[width=\textwidth]{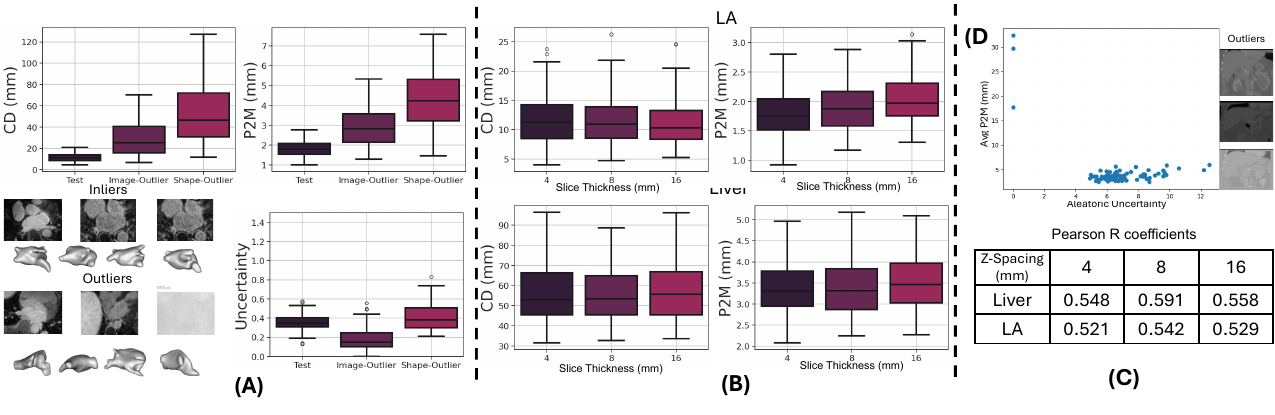}
    \vspace{-9mm}
    \caption{\textbf{Ablation Experiments:} (A) Box plots illustrate performance metrics for inliers, image outliers, and shape outliers, with example image slices. (B) Box plots compare performance metrics across varying slice thickness. (C) r-score for correlation between estimated uncertainty and P2S error for different slice thickness models. (D) Liver outlier detection results, with slices of outlier images and a plot showing total prediction uncertainty and P2S error, averaged across each shape. High-error, low-uncertainty outliers are highlighted in red. }
    \label{fig:other_exps}
    \vspace{-5mm}
\end{figure}

We experimented with the orthogonal slice dataset obtained from volumes of varying thickness levels to demonstrate the utility of using sparse images. As shown in Fig~\ref{fig:other_exps}.B, the performance metrics indicate that the model performs similarly across these versions, providing consistent aleatoric estimates, as evidenced by the r-scores in Fig~\ref{fig:other_exps}.C.

Fig~\ref{fig:p2s_ale} illustrates the point-wise correlation between predicted uncertainty values and P2S distance error across the test set. Higher uncertainty is expected for points further from the true shape surface. The Pearson R correlation coefficients show that using orthogonal images does not degrade uncertainty estimation, as indicated by the similar average uncertainty heatmaps. However, individual slices reduce uncertainty calibration due to information loss. The spatial correlation between P2S error and uncertainty heatmaps highlights the value of probabilistic frameworks in assessing prediction reliability. For the LA dataset, the correlation between P2S error and aleatoric uncertainty using the axial image is comparable to that of orthogonal and full-volume images, consistent with the SSM metrics in Fig~\ref{fig:metrics}.

Using the method described in BVIB-DeepSSM \cite{adams2023fully}, we selected outlier cases for the test set based on an outlier degree computed from images and meshes. This resulted in a test set with 40 shape outliers, 78 image outliers, and 92 randomly selected inliers. Fig~\ref{fig:other_exps}.A shows that predicted uncertainty is higher for outlier test sets, particularly extreme shape outliers, as illustrated by the examples of outliers which display high variability and differ from the inliers significantly. 

For the liver dataset, we examined the correlation between sample-wise aleatoric estimates and sample-wise P2M distance. Fig~\ref{fig:other_exps}.D highlights three samples with high P2M distance and low uncertainty, indicating confident but incorrect predictions. These errors are attributed to poor contrast and lack of clear organ definition in the image slices of these outliers, as observed in the example outlier image slices.

\vspace{-4mm}
\section{Conclusion}
\vspace{-2.5mm}
\model demonstrates substantial potential by providing a straightforward approach for directly inferring probabilistic correspondences from raw images without needing pre-optimized shape models. Leveraging shape priors from various representations and integrating aleatoric uncertainty quantification methods, \model effectively accommodates sparse images, significantly enhancing its reliability and applicability in clinical settings. The current model relies on precise image alignment for optimal performance; future work on developing robust alignment algorithms or alignment-free methods holds promise for increasing its versatility across diverse datasets and clinical scenarios. This streamlined approach to shape model generation marks a significant step forward in personalized medicine and clinical decision support, promising substantial progress and broader applicability.

\section{Acknowledgements}
This work was supported by the National Institutes of Health under grant numbers NIBIB-U24EB029011, NIAMS-R01AR076120, and NHLBI-R01HL135568. We thank the University of Utah Division of Cardiovascular Medicine for providing left atrium MRI scans and segmentations from the Atrial Fibrillation projects and the ShapeWorks team.

% \clearpage

\bibliographystyle{splncs04}
\bibliography{ref}

\begin{thebibliography}{10}
\providecommand{\url}[1]{\texttt{#1}}
\providecommand{\urlprefix}{URL }
\providecommand{\doi}[1]{https://doi.org/#1}

\bibitem{adams2020uncertain}
Adams, J., Bhalodia, R., Elhabian, S.: Uncertain-deepssm: From images to probabilistic shape models. In: Shape in Medical Imaging: International Workshop, ShapeMI 2020, Held in Conjunction with MICCAI 2020, Lima, Peru, October 4, 2020, Proceedings. pp. 57--72. Springer (2020)

\bibitem{adams2022images}
Adams, J., Elhabian, S.: From images to probabilistic anatomical shapes: A deep variational bottleneck approach. In: Medical Image Computing and Computer Assisted Intervention--MICCAI 2022: 25th International Conference, Singapore, September 18--22, 2022, Proceedings, Part II. pp. 474--484. Springer (2022)

\bibitem{adams2023point2ssm}
Adams, J., Elhabian, S.: Point2ssm: Learning morphological variations of anatomies from point cloud. arXiv preprint arXiv:2305.14486  (2023)

\bibitem{adams2023fully}
Adams, J., Elhabian, S.Y.: Fully bayesian vib-deepssm. In: International Conference on Medical Image Computing and Computer-Assisted Intervention. pp. 346--356. Springer (2023)

\bibitem{alemi2016deepvib}
Alemi, A.A., Fischer, I., Dillon, J.V., Murphy, K.: Deep variational information bottleneck. arXiv preprint arXiv:1612.00410  (2016)

\bibitem{bhalodia2024deepssm}
Bhalodia, R., Elhabian, S., Adams, J., Tao, W., Kavan, L., Whitaker, R.: Deepssm: A blueprint for image-to-shape deep learning models. Medical Image Analysis  \textbf{91},  103034 (2024)

\bibitem{cates2017shapeworks}
Cates, J., Elhabian, S., Whitaker, R.: Shapeworks: Particle-based shape correspondence and visualization software. In: Statistical Shape and Deformation Analysis, pp. 257--298. Elsevier (2017)

\bibitem{cates2008particle}
Cates, J., Fletcher, P.T., Styner, M., Hazlett, H.C., Whitaker, R.: Particle-based shape analysis of multi-object complexes. In: International Conference on Medical Image Computing and Computer-Assisted Intervention. pp. 477--485. Springer (2008)

\bibitem{chen2019net}
Chen, Z.: Im-net: Learning implicit fields for generative shape modeling  (2019)

\bibitem{dai2020statistical}
Dai, H., Pears, N., Smith, W., Duncan, C.: Statistical modeling of craniofacial shape and texture. International Journal of Computer Vision  \textbf{128}(2),  547--571 (2020)

\bibitem{davies2002learning}
Davies, R.H.: Learning shape: optimal models for analysing natural variability. The University of Manchester (United Kingdom) (2002)

\bibitem{durrleman2014morphometry}
Durrleman, S., Prastawa, M., Charon, N., Korenberg, J.R., Joshi, S., Gerig, G., Trouv{\'e}, A.: Morphometry of anatomical shape complexes with dense deformations and sparse parameters. NeuroImage  \textbf{101},  35--49 (2014)

\bibitem{gal2016uncertainty}
Gal, Y., et~al.: Uncertainty in deep learning  (2016)

\bibitem{heitz2005statistical}
Heitz, G., Rohlfing, T., Maurer~Jr, C.R.: Statistical shape model generation using nonrigid deformation of a template mesh. In: Medical Imaging 2005: Image Processing. vol.~5747, pp. 1411--1421. SPIE (2005)

\bibitem{hollingsworth2015reducing}
Hollingsworth, K.G.: Reducing acquisition time in clinical mri by data undersampling and compressed sensing reconstruction. Physics in Medicine \& Biology  \textbf{60}(21), ~R297 (2015)

\bibitem{iyer2024scorp}
Iyer, K., Adams, J., Elhabian, S.Y.: Scorp: Statistics-informed dense correspondence prediction directly from unsegmented medical images. arXiv preprint arXiv:2404.17967  (2024)

\bibitem{iyer2023mesh2ssm}
Iyer, K., Elhabian, S.Y.: Mesh2ssm: From surface meshes to statistical shape models of anatomy. In: International Conference on Medical Image Computing and Computer-Assisted Intervention. pp. 615--625. Springer (2023)

\bibitem{li2024back}
Li, J., Ellis, D.G., Pepe, A., Gsaxner, C., Aizenberg, M.R., Kleesiek, J., Egger, J.: Back to the roots: Reconstructing large and complex cranial defects using an image-based statistical shape model. Journal of Medical Systems  \textbf{48}(1), ~55 (2024)

\bibitem{ludke2022landmark}
L{\"u}dke, D., Amiranashvili, T., Ambellan, F., Ezhov, I., Menze, B.H., Zachow, S.: Landmark-free statistical shape modeling via neural flow deformations. In: Medical Image Computing and Computer Assisted Intervention--MICCAI 2022: 25th International Conference, Singapore, September 18--22, 2022, Proceedings, Part II. pp. 453--463. Springer (2022)

\bibitem{Ma-2021-AbdomenCT-1K}
Ma, J., Zhang, Y., Gu, S., Zhu, C., Ge, C., Zhang, Y., An, X., Wang, C., Wang, Q., Liu, X., Cao, S., Zhang, Q., Liu, S., Wang, Y., Li, Y., He, J., Yang, X.: Abdomenct-1k: Is abdominal organ segmentation a solved problem? IEEE Transactions on Pattern Analysis and Machine Intelligence  \textbf{44}(10),  6695--6714 (2022). \doi{10.1109/TPAMI.2021.3100536}

\bibitem{munsell2008evaluating}
Munsell, B.C., Dalal, P., Wang, S.: Evaluating shape correspondence for statistical shape analysis: A benchmark study. IEEE Transactions on Pattern Analysis and Machine Intelligence  \textbf{30}(11),  2023--2039 (2008)

\bibitem{paulsen2002building}
Paulsen, R., Larsen, R., Nielsen, C., Laugesen, S., Ersb{\o}ll, B.: Building and testing a statistical shape model of the human ear canal. In: International Conference on Medical Image Computing and Computer-Assisted Intervention. pp. 373--380. Springer (2002)

\bibitem{peirlinck2021precision}
Peirlinck, M., Costabal, F.S., Yao, J., Guccione, J., Tripathy, S., Wang, Y., Ozturk, D., Segars, P., Morrison, T., Levine, S., et~al.: Precision medicine in human heart modeling: Perspectives, challenges, and opportunities. Biomechanics and modeling in mechanobiology  \textbf{20},  803--831 (2021)

\bibitem{samson2000level}
Samson, C., Blanc-F{\'e}raud, L., Aubert, G., Zerubia, J.: A level set model for image classification. International journal of computer vision  \textbf{40}(3),  187--197 (2000)

\bibitem{schultz2020risk}
Schultz, C.H., Fairley, R., Murphy, L.S.L., Doss, M.: The risk of cancer from ct scans and other sources of low-dose radiation: a critical appraisal of methodologic quality. Prehospital and disaster medicine  \textbf{35}(1),  3--16 (2020)

\bibitem{shieh2019spare}
Shieh, C.C., Gonzalez, Y., Li, B., Jia, X., Rit, S., Mory, C., Riblett, M., Hugo, G., Zhang, Y., Jiang, Z., et~al.: Spare: Sparse-view reconstruction challenge for 4d cone-beam ct from a 1-min scan. Medical physics  \textbf{46}(9),  3799--3811 (2019)

\bibitem{singh2020evaluation}
Singh, B., Kumar, N.R., Balan, A., Nishan, M., Haris, P., Jinisha, M., Denny, C.D.: Evaluation of normal morphology of mandibular condyle: a radiographic survey. Journal of clinical imaging science  \textbf{10} (2020)

\bibitem{styner2006framework}
Styner, M., Oguz, I., Xu, S., Brechb{\"u}hler, C., Pantazis, D., Levitt, J.J., Shenton, M.E., Gerig, G.: Framework for the statistical shape analysis of brain structures using spharm-pdm. The insight journal (1071), ~242 (2006)

\bibitem{tothova2020probabilistic}
T{\'o}thov{\'a}, K., Parisot, S., Lee, M., Puyol-Ant{\'o}n, E., King, A., Pollefeys, M., Konukoglu, E.: Probabilistic 3d surface reconstruction from sparse mri information. In: Medical Image Computing and Computer Assisted Intervention--MICCAI 2020: 23rd International Conference, Lima, Peru, October 4--8, 2020, Proceedings, Part I 23. pp. 813--823. Springer (2020)

\bibitem{tothova2018uncertainty}
T{\'o}thov{\'a}, K., Parisot, S., Lee, M.C., Puyol-Ant{\'o}n, E., Koch, L.M., King, A.P., Konukoglu, E., Pollefeys, M.: Uncertainty quantification in cnn-based surface prediction using shape priors. In: Shape in Medical Imaging: International Workshop, ShapeMI 2018, Held in Conjunction with MICCAI 2018, Granada, Spain, September 20, 2018, Proceedings. pp. 300--310. Springer (2018)

\bibitem{wang2019dynamic}
Wang, Y., Sun, Y., Liu, Z., Sarma, S.E., Bronstein, M.M., Solomon, J.M.: Dynamic graph cnn for learning on point clouds. Acm Transactions On Graphics (tog)  \textbf{38}(5),  1--12 (2019)

\end{thebibliography}

\appendix
\section{Appendix}
\subsection{Dataset Details}
\begin{itemize}
    \item Left Atrium (LA)
    
    \begin{itemize}
        \item 923 anonymized Late Gadolinium Enhancement (LGE) MRIs from distinct patients.
        \item Manually segmented by cardiovascular medicine experts at the (anonymous) Cardiovascular Medicine.
        \item The endocardial wall was used to cut off pulmonary veins.
        \item Spatial resolution: \(0.65 \times 0.65 \times 2.5 \ \mathrm{mm}^3\).
        \item Images were cropped around the region of interest and downsampled by a factor of 0.8.
        \item Resulting input image size: \(166 \times 120 \times 125\).
    \end{itemize}

    \item Liver 
    
    \begin{itemize}
        \item Dataset includes CT scans and segmentations of liver, kidney, spleen, and pancreas.
        \item 1132 3D CT scans from various public datasets with segmentation verified and refined by experienced radiologists.
        \item Used CT scans and corresponding liver segmentations for experiments.
        \item CT scans have resolutions of \(512 \times 512\) pixels with varying pixel sizes and slice thicknesses between 1.25-5 mm.
        \item Utilized 833 samples after visual quality assessment of images and segmentations.
        \item Images were cropped around the region of interest using segmentations and downsampled by a factor of 3.5.
        \item Downsampled volume size: \(144 \times 156 \times 115\) with isotropic voxel spacing of 2 mm.
    \end{itemize}

\end{itemize}

\subsection{Hyperparamters}
All models were trained on NVIDIA GeForce RTX 2080 Ti. 
\begin{table}[ht]
    \centering
    \begin{tabular}{|c|c|c|}
    \hline
    Parameter & Description & Value\\
    \hline
        B & Batch size & 6  \\
        LR & Learning rate & \(1e^{-5}\)\\
        M & Number of correspondences & 1024 \\
        ES & Early stopping patience epochs & 200 \\
        L & Latent dimension for \model & 256 \\
        K & Size of neighbourhood for EdgeConv & 27\\    
        NV & Number of vertices in the mesh & 5000 \\

    \hline
    \end{tabular}
    \caption{Hyperparameters for \model}
    \label{tab:hyperparameters}
\end{table}

\subsection{Architecture}
\begin{enumerate}
    \item \textbf{Orthogonal Encoder:} The Orthogonal Encoder processes three orthogonal 2D slices (axial, sagittal, and coronal) from a 3D medical image volume.

        \begin{itemize}
            \item \textbf{Slice Encoders:} Separate 2D convolutional backbones are used for each of the three slices: axial, sagittal, and coronal. Each backbone processes its respective slice using Conv2d layers with \(5\times 5\) filters and the following numbers of filters: \([12, 24, 48, 96, 192]\). Batch normalization and ReLU activation functions are applied after each Conv2d layer, with max-pooling layers incorporated to reduce spatial dimensions.
        
            \item \textbf{Fully Connected Layer:} The combined features are passed through a fully connected (FC) layer stack. This stack includes two linear layers: \([256 \times 3 \rightarrow 256]\) and \([256 \rightarrow \texttt{z\_dim}]\), with a Parametric ReLU (PReLU) activation function in between.
            \item \textbf{Output:} 
            \begin{itemize}
                \item If the encoder is deterministic, the output is directly the features from the FC layer.
                \item If the encoder is non-deterministic, the output is split into mean and log variance for Gaussian sampling, producing the required number of samples.
            \end{itemize}
        \end{itemize}

    \item \textbf{3D Image encoder:} The encoder architecture utilizes Conv2d layers with \(5\times 5\) filters and the following numbers of filters: \([12, 24, 48, 96, 192]\). After each Conv2d layer, batch normalization and ReLU activation functions are applied. Max pooling layers are incorporated to reduce spatial dimensions. The feature maps are then flattened and passed to the fully connected layers. The fully connected (FC) layer stack consists of linear layers with different input and output feature dimensions: \([193536 -> 384], [384 -> 96], [96 -> 256]\). Each linear layer is followed by a Parametric ReLU (PReLU) activation function. 
    \item \textbf{2D Orthogonal Slice Image encoder:}
    \item \textbf{Image Feature Aggregator:}
    \item \textbf{Surface Autoencoder: } We use the \textit{DGCNN\_semseg\_s3dis} model from the original DGCNN \href{https://github.com/antao97/dgcnn.pytorch/}{Github} repository. 
    \item \textbf{IM-Net:} We use the original implementation of IM-Net from the \href{https://github.com/czq142857/IM-NET-pytorch}{Github} repository. 
\end{enumerate}
\subsection{SSM Metrics}
\begin{enumerate}
    \item Compactness:  We quantify compactness as the number of
PCA modes that are required to capture \(95\%\) of the total variation in the output training cohort correspondence points.
    \item Specificity: We quantify specificity by randomly generating \(J\) samples from the shape space using the eigenvectors and eigenvalues that capture \(95\%\) variability of the training cohort. Specificity is computed as the average squared Euclidean distance between these generated samples and their closest training sample.\\
    \( S= \sum_{\set{C} \in \set{C}_{generated}} ||\set{C} - \set{C}_{train}||^2 \)\\

    \item Generalization: We quantify generalization by assessing the average approximation errors across a set of unseen instances. Generalization is defined as the mean approximation errors between the original unseen shape instance and reconstruction of the shape constructed using the raining cohort PCA eigenvalues and vectors that preserve \(95\%\) variability. \\
    \( G = \sum_{j=1}^U||\set{C}_j - \hat{\set{C}_{j}}||_2^2\) for J unseen shapes.

\end{enumerate}
\end{document}